# EMOTION DETECTION WITH TRANSFORMERS: A COMPARATIVE STUDY

**Mahdi Rezapour**

## ABSTRACT

In this study, we explore the application of transformer-based models for emotion classification on text data. We train and evaluate several pre-trained transformer models, on the Emotion dataset using different variants of transformers, such as DistilBERT and a recurrent neural network layer of long short-term memory (LSTM) with Global vectors for word representation (GloVe) embeddings, to capture the semantic and emotional features of the text. The experimental results show that although all models perform very closely, the pre-trained architecture of twitter-roberta-base slightly outperformed other models with an accuracy of 92% despite the relatively small number of observations available for training. The paper also analyzes some factors that influence the performance of the model, such as the fine-tuning of the transformer layer, the trainability of the layer, and the preprocessing of the text data. Our analysis reveals that commonly applied techniques like removing punctuation and stopwords can hinder model performance. This might be because transformers' strength lies in understanding contextual relationships within text. Elements like punctuation and stopwords can still convey sentiment or emphasis and removing them might disrupt this context.

## 1 Introduction

Social media platforms are widely used by people to share their opinions and emotions about various topics. Analyzing the sentiment and emotion behind social media text data can help us understand the attitudes, preferences, and feelings of the users [12]. Unveiling these emotions goes beyond simply gauging sentiment, as it provides deeper insights into user motivations and psychological states. For instance, capturing the nuances of human emotion expressed through text remains a complex task due to the limitations of language itself. Sentiment analysis, while valuable, only provides a surface-level understanding. Identifying specific emotions expressed in text offers deeper insights into user behavior and motivations.

Sentiment analysis is a natural language processing task that aims to classify the polarity of a text as positive, negative, or neutral [7] [4]. Emotion classification is a related task that aims to identify the specific emotion expressed in a text, such as sadness, joy, love, anger, fear, or surprise. Both tasks might be challenging due to the complexity and variability of natural language.

Transfer learning is a technique that allows a model trained on one task to be used for another related task [19]. For instance, the use of transfer learning in sentiment analysis using different transformer models including Bidirectional Encoder Representations from Transformers (BERT), financial version of BERT (finBERT), eXtreme Multi-Lingual Language Understanding (XLNet), Cross-lingual Language Model. (XLM) , a Lite BERT (ALBERT), A Robustly Optimized BERT Pretraining Approach (RoBERTa), a distillation BERT (DistlBERT), XLM-RoBERTa and Bidirectional Autoregressive Transfer Transformer (BART) have been employed in past studies e.g., [18].

These transformer models have been pre-trained on large amounts of unlabeled data and fine-tuned on labeled data for specific tasks. They have achieved state-of-the-art performance on various natural language processing tasks such as text classification, named entity recognition, and question answering.

Those attention networks could be used to alleviate the issues associated with traditional models like Recurrent Neural Network (RNN). The goal of those model is to focus on part of text that are of high interest [10]. [22] proposed the transformer single structure model that is equipped with attention structure. Those attention layers are based on gaining knowledge from one task and applying that to other related tasks. Paul Ekman's model categorizes emotions into six categories, including happiness, sadness, anger. Disgust, surprise and fear [20]. They classify the emotions of utterances from Twitter conversations as happy, sad, angry and others [5]. They used ensemble approach model made up of a Hierarchical LSTM for emotion detection and BERT model.

A deep learning approach with BERT was used to detect cyber abuse in English and Hindi texts, [2]. In that study, the texts were classified into 3 categories: overly, covertly and non -aggressive. A framework was proposed to identify dimensional valence arousal dimension scores from the corpus with categorical emotion labels [24]. They utilized a fine-tuned, pre-trained BERT model.

Suicide risks was assessed using multi-level language using BERT [3]. The BAKE and exBAKE models were designed to automatically detect fake new using BERT through scrutiny of headlines and body texts of news [13]. The efficacy of BERT, RoBERTa, DIStilBERT, and XLNet transformers models were used in recognizing emotions from texts. The model could distinguish emotions into anger, disgust, sadness, fear, joy, shame and guilt [1].

variant of BERT, SentiBERT, were used, which effectively captures compositional sentiment semantics [8]. SentiBERT incorporates contextualized representation with a binary constituency parse tree to capture semantic composition. The binary constituency parse tree was used to generate auxiliary sentences that represent the sentiment polarity of each phrase in the main sentence. The experiments demonstrated that SentiBERT achieved competitive performance on phrase-level sentiment classification. They also showed that the sentiment composition that was learned from the phrase-level annotations on SST was shown to be transferable to other sentiment analysis tasks, as well as related tasks such as emotion classification.

In this paper, we will explore the use of transformers in sentiment analysis. Transformers are a type of deep learning model that uses self-attention to differentially weight the significance of each part of the input data [22]. They are used in computer vision (CV) and natural language processing (NLP). The transformer architecture aims to solve sequence-to-sequence tasks, while handling long-range dependencies with ease (Zheng et al., 2021).

## 2 Method

In this section, we present the methods that we applied to fine-tune different transformer models for emotion classification. We selected these methods because they are based on BERT, a powerful and widely-used language model for natural language processing tasks. However, instead of using BERT directly, we used two of its variants: distilbert-base-uncased and electra-base-discriminator.

We also used twitter-roberta-base-emotion, which is a pre-trained model specifically designed for emotion classification on Twitter data. Additionally, we used LawalAfeez/emotion_detection (https://huggingface.co/LawalAfeez/emotion_detection/tree/main, 2021), and finally employed long short-term memory (LSTM) with GloVe embedding to capture the contextual and semantic features of the text. It should be noted that "LawalAfeez/emotion_detection" model is a fine-tuned version of distilbert-base-uncased hosted on the Hugging Face platform. It has been specifically designed for emotion detection tasks.

### 2.1 BERT

BERT is a large language model that was developed by Google AI. BERT is a powerful tool for natural language processing tasks, such as text classification, sentiment analysis, and question answering. The techniques that we used in my paper are based on BERT. However, we did not use BERT directly. Instead, we used two variants of BERT: distilbert-base-uncased and electra-base-discriminator.

Depending on the task and resources, various types of Bert model might be used. DistilBERT-base-multilingual-cased is a distilled version of the BERT base multilingual model. It is trained on the concatenation of Wikipedia in 104 different languages [26]. The model has 6 layers, 768 dimensions and 12 heads, attention heads, totaling 134M parameters, where heads are components of the transformer models implementing the self-attention mechanisms.

Before feeding our own model to the pre-trained model, we need to transform the text to the same structure as the pre-trained model so it can understand it. This means that we need to use the same tokenizer and vocabulary as the pre-trained model to convert your text into a sequence of tokens. The pre-trained model does not have any text in the sense of sentences or paragraphs, but it has a vocabulary in the sense of tokens or words.

BERT takes an input of a concatenation of two sequences of token, $x_1, \ldots, x_N$ and second segment sequence of $y_1, \ldots, y_N$, where $M + N < T$, and those two segments would be presented as a single input sequence as $[CLS], x_1, \ldots, x_N, [SEP], y_1, \ldots, y_M, [EOS]$. EOS is a special token that is used to indicate the end of a sequence of tokens, while classification (CLS) is used as the first token in the input sequence, and [SEP] donates the end of a sentence.

BERT used two objectives, masked language modeling and next-sentence prediction. Masked languages model (MLM) includes randomly masking tokens of the input sequence and replace them with [MASK], while using cross-entropy loss on predicting the masked tokens. BERT chooses 15% of the input tokens for replacement, where 50% of those are replaced with [MASK], and 10% are left unchanged, and 10% are replaced by randomly selected vocabularies.

On the other hand, the next-sentence prediction (NSP) uses a binary classification loss, checking whether two segments follow each other in the original text or not. BERT is trained on a combination of English Wikipedia and BookCorpus, totaling 16GB of uncompressed text [28].

BERT multi-layer bidirectional Transformer, on the other hand, is trained on next-sentence prediction tasks and plain text for masked word prediction. Unlike bidirectional language models (biLM), being limited to a combination of two unidirectional language models (i.e., right-to-left and left-to-right), BERT uses a Masked Language. Different levels of semantic and syntactic information were captured by different layers of BERT [15].

### 2.2 distilbert-base-uncased

This method was proposed to pretrain a smaller general-purpose language representation model called DistilBERT, being fine-tuned with good performance on a wide range of tasks like its larger counterparts [26]. It was shown that they could reduce the BERT model by 40%, while retaining 97% of its language understanding capabilities and being 60% faster. The triple loss, which comprises language modeling, distillation, and cosine-distance losses, was intentionally introduced to harness the inductive biases acquired by larger models during pretraining. These biases serve as valuable priors that enhance the performance of the distilled model, DistilBERT. Here, the distillation loss was used to transfer knowledge from a larger model to a smaller model, where cosine distance loss minimizes the distance between the embeddings of the original and the distilled models.

Consideration of this mode, compared with BERT, is especially important due to the main concern of computational requirements of these types of model [23]. Knowledge distillation is a technique used to compress a larger model, known as the teacher, into a smaller model, known as the student. The student model is trained to mimic the behavior of the teacher model or an ensemble of models. [11]. It was discussed that they can transfer knowledge from the cumbersome model to a smaller model that is more suitable for deployment (Distilling the Knowledge in a Neural Network).

In the case of DistilBERT, the KL divergence loss function is used to calculate the distillation loss between the outputs of the teacher (BERT) and student (DistilBERT) models. This distillation loss is used in addition to the cross-entropy loss to train the student model. In the case of BERT and DistilBERT, both models are trained on the same task and use the same dataset. The difference is that DistilBERT is a smaller and simpler model that has been trained using knowledge distillation from the larger and more complex BERT model. BERT is a larger and more complex model that has 340 million parameters, while DistilBERT has only 66 million parameters. The cross-entropy loss is used to train the student model on the true labels, while the KLD loss is used to train the student model on the soft targets generated by the teacher model.

### 2.3 electra-base-discriminator

ELECTRA, or Efficiently Learning an Encoder that Classifies Token Replacement Accurately, is the name of a method used to learn an encoder. It was found that ELECTRA-large model performs comparably to RoBERTa [27] and XLNet [29].

ELECTRA is a variant of BERT, meaning that they have similar layers, while they have different pre-training methods. ELECTRA used the same data as BERT, consisting of 3.3 billion tokens from Wikipedia and BookCorpus [17]. In this method, the model learns to distinguish real input tokens from plausible but synthetically generated replacements [17]. That is done by corrupting the output of a small, masked language model, instead of masking.

The model has two transformers' models: the generator and the discriminator. The generator acts like a masked language model, while the discriminator tries to identify which tokens were replaced by the generator in the sequence. This method addresses the computational costs of BERT, which requires a large amount of compute to be effective.

This technique, instead of masking the input, corrupts it by replacing some tokens with plausible alternatives. These alternatives are sampled from a small generator network.

As discussed, the model consists of Masked language modeling (MLM) and a discriminator. The process includes two neural networks, a generator $G$ and a discriminator $D$. For a given position $t$, the generator outputs a probability for generating a particular token $x_t = [MASK]$, with a SoftMax layer as:

$$P_G(x_t|x) = exp(e(x_t)^T h_G(x)_t) / \sum_{x'} exp(e(x'_t)^T h_G(x)_t) \quad (1)$$

where encoder (e.g., a transformer network) maps a sequence of input tokens $x=[x_1, \ldots, x_n]$ into a sequence of contextualized vector representations $h(x) = [h_1, \ldots, h_n]$. The above probability is for a position $t$, $x_t = [MASK]$, and $e$ donates token embeddings. Now the discriminator predicts whether the token $x_t$ is 'real', coming from data rather than the generator distribution, as

$$D(x, t) = sigmoid(w^T h_D(x)_t) \quad (2)$$

Initially google/electra-small-discriminator and then google/electra-base-discriminator were used [17]. google/electra-small-discriminator is a small version of the ELECTRA model. It is a transformer-based neural network that can be used for natural language processing tasks such as text classification, question answering, and sequence tagging.

ELECTRA has several advantages over other pre-training methods such as BERT. For instance, it is more sample-efficient, indicating that it could achieve better results with less data, and it is more computationally efficient, it can be trained faster and with smaller models, also it has a better performance on downstream tasks, especially on small datasets.

ELECTRA-small has 4 attention heads, 256 hidden size, 128 embedding size and 12 hidden layers, while ELECTRA-base has 12 attention heads, 768 hidden size, 128 embedding size, and 12 hidden layers [17]. Electra-base discriminator has similar architecture but higher number of parameters, 137M versus 110M parameters for ELECTRA-base. The main difference between ELECTRA-base and ELECTRA-base-discriminator is the training objective. ELECTRA-base is trained using a masked language modeling objective, while ELECTRA-base-discriminator is trained using a discriminator objective.

### 2.4 LawalAfeez/emotion$_d$etection

This model is a fine-tuned version of distilbert-base-uncased on an unknown dataset, where that was fine-tuned specifically for the task of emotion detection in text data. This fine-tuning involves training the pre-trained DistilBERT model on a labeled emotion detection dataset. In summary, in the case of LawalAfeez/emotion_detection, it is a fine-tuned version of distilbert-base-uncased. In this model, the pre-trained model of distilbert is further trained on a specific task. That model used optimizerAdam, with learning_rate of 5e-05, 'decay': 0.0, 'beta_1': 0.9, 'beta_2': 0.999, 'epsilon': 1e-07, 'amsgrad': False.

### 2.5 twitter-roberta-base-emotion

Robustly optimized BERT approach (RoBERTa) is a robustly optimized BERT approach that exceeds the original BERT model performance. The changes that were made to RoBERTa include: model training for longer periods of time, with larger batches, on more data, removing the next sentence prediction objective, training on longer sequences, and dynamically changing the masking pattern applied to the training data [27].

The RoBERTa-base model was trained on about 58M tweets and fine-tuned for emotion recognition with the TweetEval benchmark. The TweetEval benchmark is a new evaluation framework for Twitter-specific classification tasks, including sentiment analysis, emotion recognition, offensive language detection, emoji prediction, and irony detection.

The RoBERTa model has the same architecture as BERT, but it uses a byte-level byte pair encoding (BPE) tokenizer. BPE is a data compression technique that replaces the most frequent pairs of bytes in a sequence with a single, unused byte [14]. The byte-level BPE is a variant of BPE that operates on bytes instead of characters. The BPE tokenizer is more efficient and can handle out-of-vocabulary words better than the WordPiece tokenizer.

In terms of next sentence prediction, RoBERTa does not use this task during pre-training. Instead, it uses a larger corpus of text and a longer training time to improve the performance of the model. The model was trained on five

English-language corpora of varying sizes, totaling over 160GB of uncompressed text. The text includes various datasets such as BOOKCORPUS, CC-NEWS, OPENWEBTEXT, and STORIES.

RoBERTa was evaluated on downstream tasks using the three benchmarks of the General Language Understanding Evaluation (GLUE), the Reading Comprehension from Examination (RACE) and the Stanford Question Answering Dataset (SQuAD). We know, the GLUE benchmark is a collection of 9 datasets for evaluation of natural language understanding systems. SQuAD is a Stanford Question Answering Dataset that is used to evaluate the ability of a model to answer questions by extracting the relevant span from the context [21]. On the other hand, RACE is a large-scale reading comprehension dataset with more than 28,000 passages and nearly 100,000 questions.

The authors of the RoBERTa paper discuss that while BERT relies on random masking and predicting tokens once during data processing to avoid using the same mask, in every epoch, the training data was duplicated 10 times and thus the sequence was masked in 10 different ways over all 40 epochs. This is called dynamic masking. [27]. The results of the experiments showed that RoBERTa outperforms BERT on all three benchmarks. This suggests that RoBERTa is a more effective model for natural language understanding tasks.

## 2.6 LSTM with GloVe pretraining

LSTM is a type of neural network that can process sequential data, such as text. GloVe is a method that creates word embeddings, which are numerical representations of words that capture their meanings and contexts. LSTM can use GloVe embeddings as input to learn from pre-trained word vectors. LSTM is a supervised learning model, which means it learns from data that has labels such as the emotion of a sentence.

The model is trained using the categorical_crossentropy loss function, which measures the difference between the predicted probability distribution and the true probability distribution over the classes. The Adam optimizer is used to minimize this loss function during training. The categorical_accuracy metric is employed, on the other hand, to evaluate the performance of the model during training and testing. During training, the model updates its weights based on the gradients of the loss function with respect to its parameters using backpropagation through time (BPTT). The dropout layers are used to prevent overfitting by randomly dropping out some of the units during training.

Our LSTM layer has 100 units and a dropout of 0.3. The model has a dense layer with 4 units and uses the softmax activation function. The model is compiled with the categorical_crossentropy loss function, the Adam optimizer, and the categorical_accuracy metric.

Word embeddings are a type of pre-trained model that maps words to vectors of real numbers, where these vectors can be used as features in machine learning models for natural language processing tasks such as text classification, sentiment analysis, and named entity recognition. Word2vec [25] and GloVe [16] are two popular algorithms for generating word embeddings.

GloVe is a count-based algorithm that learns word embeddings by factorizing a matrix of word co-occurrence statistics. It is an extension of the word2vec algorithm, where Word2vec is a neural network-based algorithm that learns word embeddings by predicting the context of words in a large corpus of text, that creates the Global Vectors for Word Representation algorithm.

The embedding matrix is created by mapping each word in the dataset to its corresponding embedding vector in the dictionary. The LSTM layer processes the sequence of embeddings and outputs a sequence of hidden states. The dense layer takes the final hidden state of the LSTM layer as input and outputs a probability distribution over the classes.

However, compared with Glove, BERT, for instance, uses different word embedding methods using bidirectional transformers to process text. Bidirectional encoder representations from transformers (BERT) expect the input data in specific format that includes special tokens and word piece tokenization, so BERT relies on full context of the text, including stopwords for capturing the meaning and thus to perform well.

For instance, GloVe uses a dot product to measure the similarity between words, and it can produce word analogies, such as king+woman-man=queen. BERT uses a masked language model and next sentence prediction objective to learn from the context of words and sentences.

## 2.7 study objectives

The main objective of this study is to compare different methods of fine-tuning transformer models for emotion classification. Emotion classification is a fine-grained text classification task that requires distinguishing subtle differences between similar or overlapping emotion categories. Transformer models, such as BERT, have been shown to achieve high performance on text classification tasks by leveraging large-scale pre-trained language representations.

However, fine-tuning transformer models for specific tasks and domains may pose some challenges, such as data scarcity, class imbalance, and domain adaptation. Therefore, we conducted extensive experiments using different techniques to fine-tune various transformer models for emotion classification on a large and diverse dataset of Reddit comments.

## 2.8 Data

We used the GoEmotions dataset as our data source for emotion classification. The GoEmotions dataset is a corpus of 58k Reddit comments, labeled for 27 emotion categories or Neutral, created by Google Research [6]. It is the largest manually annotated dataset of fine-grained emotions in English to date, and it covers a wide range of positive, negative, and ambiguous emotions. The dataset also provides information about the context and the target of each comment.

For the purpose of this study, we only considered four emotions: Anger, Admiration, Amusement, and Love. We also excluded those comments that had multiple or conflicting emotion labels. We ended up with a total of 8,455 comments for our emotion classification task. We randomly split the data into 70% for training (5,918 comments) and 30% for testing (2,537 comments). The training data was imbalanced, with Amusement being the most frequent emotion (40%) and Anger being the least frequent emotion (15%).

## 2.9 Text preprocessing

Text preprocessing is an important step in NLP that involves cleaning and transforming raw text data into a format that can be used by machine learning models. The goal of text preprocessing is to remove irrelevant information from the text data and transform it into a format that can be easily analyzed by machine learning algorithms.

The preprocessing task removes HTML tags, punctuation characters, digits, links and URLs, special characters, stopwords, non-ASCII characters, and email addresses from the text data. These are common text preprocessing steps that might help improve the performance of machine learning models on NLP tasks.

It's worth noting that the specific text preprocessing steps we choose to perform may depend on the specific NLP task we are working on. For example, some NLP tasks may require keeping certain types of punctuation or special characters in the text data. Similarly, some NLP tasks may require keeping certain stop words in the text data. Therefore, it's important to carefully consider which text preprocessing steps are appropriate for your specific NLP task.

While text preprocessing is an important step in NLP, it's worth noting that some machine learning models like BERT are designed to work well with raw text data without any preprocessing. In fact, some studies have shown that text preprocessing can actually hurt the performance of BERT on certain NLP tasks [9]. Therefore, it's important to carefully consider whether or not text preprocessing is necessary for your specific NLP task when using models like BERT.

## 3 Results

We evaluated the performance of different models on the emotion classification task using four metrics: recall, precision, F1, and accuracy. Table 1 shows the confusion matrix and the evaluation metrics for each model on the test set. The confusion matrix shows the number of predictions for each emotion class, as well as the number of correct and incorrect predictions. The best performing model on the test set was twitter-roberta-base-emotion with preprocessing, achieving an accuracy of 92%. This model outperformed all other models on all metrics, except for precision, where it was slightly lower than electra-base-discriminator with no preprocessing. This suggests that this model was able to correctly classify most of the emotions in the test set, with few errors.

Some of the key points that we considered while training these models are as follows:

Tokenization: We used the tokenizer provided by different libraries to convert each word in the text data to a unique ID that could be used as input to the model.

Padding and truncation: We padded or truncated the text data to ensure that all sequences were the same length. This is necessary because transformers require fixed-length inputs.

Fine-tuning: We fine-tuned the models on the Emotion dataset using backpropagation and gradient descent. This allowed the models to learn task-specific features that are important for emotion classification.

Regularization: We regularized the models using techniques such as dropout and batch normalization to prevent overfitting and improve generalization performance.

Optimization: We optimized the models using the Adam optimizer with a learning rate of 5e-5, epsilon of 2e-8, weight decay of 1e-2, and clipnorm of 1.0.

add_special_tokens=True: We set this parameter to True when tokenizing the text data using transformers. This tells the tokenizer to add special tokens such as [CLS] and [SEP] to the beginning and end of each sequence. These tokens are used by transformers to perform classification tasks.

return_attention_mask=True: We also set this parameter to True when tokenizing the text data. It tells the tokenizer to return an attention mask along with the input IDs. The attention mask is a binary tensor that indicates which tokens should be attended to by transformers and which should be ignored. The attention mask tensor is used to mask out padding tokens so that they are not attended to by the model. The transformers like bert_model, return a tuple containing two tensors: the final hidden state and the pooling output. The final hidden state is a tensor of shape (batch_size, sequence_length, hidden_size) that contains the final hidden states of all tokens in the input sequences. The pooling output is a tensor of shape (batch_size, hidden_size) that contains a summary of the final hidden states.

| | | Anger | Admiration | Amusement | Love | recall | precision | F1 | Accuracy |
|---|---|---|---|---|---|---|---|---|---|
| twitter-roberta-base-emotion, processing | Anger | 337 | 21 | 11 | 11 | 91% | 91% | 91% | 91% |
| | Admiration | 15 | 938 | 22 | 40 | | | | |
| | Amusement | 10 | 29 | 570 | 5 | | | | |
| | Love | 8 | 50 | 5 | 465 | | | | |
| LawalAfeez/emotion_detection, Processing | Anger | 323 | 26 | 19 | 12 | 89% | 89% | 89% | 89% |
| | Admiration | 27 | 898 | 30 | 60 | | | | |
| | Amusement | 11 | 18 | 580 | 5 | | | | |
| | Love | 8 | 33 | 25 | 462 | | | | |
| distilbert-base-uncased, Processing | Anger | 440 | 26 | 56 | 6 | 88% | 90% | 88% | 89% |
| | Admiration | 2 | 552 | 33 | 7 | | | | |
| | Amusement | 34 | 27 | 940 | 14 | | | | |
| | Love | 10 | 33 | 29 | 308 | | | | |
| twitter-roberta-base-emotion, with no preprocessing | Anger | 593 | 6 | 4 | 11 | 91% | 91% | 92% | 92% |
| | Admiration | 6 | 503 | 19 | 0 | | | | |
| | Amusement | 46 | 63 | 882 | 24 | | | | |
| | Love | 18 | 1 | 5 | 356 | | | | |
| electra-base-discriminator, with no processing | Anger | 427 | 95 | 4 | 1 | 89% | 92% | 90% | 91% |
| | Admiration | 20 | 586 | 7 | 1 | | | | |
| | Amusement | 4 | 25 | 957 | 29 | | | | |
| | Love | 22 | 27 | 4 | 327 | | | | |
| LSTM with GloVe embedding | Anger | 458 | 47 | 9 | 14 | 90% | 90% | 90% | 90% |
| | Admiration | 24 | 564 | 23 | 3 | | | | |
| | Amusement | 36 | 25 | 930 | 24 | | | | |
| | Love | 27 | 18 | 5 | 338 | | | | |

Table 1: Comparison of different methods for emotion classification on the GoEmotions dataset, test dataset.

## 4 Discussion

In this study, we compared the performance of different transformer models and LSTM with glove embedding on the emotion detection task. We used the GoEmotions dataset, which contains labeled sentences. We found that the twitter-roberta-base-emotion model, which was trained on a dataset of tweets with auxiliary sentences, achieved the best results on the test set, with an accuracy of 92%. This model outperformed all other models on all metrics, except for precision, where it was slightly lower than electra-base-discriminator. We also found that preprocessing the data did not improve model performance and led to a loss of accuracy. This suggests that the transformers can handle raw data without any modifications.

Our study has some limitations that should be considered in future work. First, we used a relatively small and imbalanced dataset, which may limit the generalizability of our findings. We acknowledge the imbalanced nature of the GoEmotions dataset, as some emotions are more frequent than others. However, during the evaluation of our models, we did not observe any significant bias towards the majority class. Therefore, we opted against data imputation or class weighting techniques in this instance.

Second, the study only evaluated the model performance on the GoEmotions dataset, which may limit the generalizability of the findings. To address this limitation, future work could involve evaluating the models on other emotion recognition datasets or even in different application domains. This broader evaluation would provide a more comprehensive understanding of the models' capabilities and limitations across various contexts.

Third, we only experimented with four transformer models and one LSTM model, which may not cover the full range of possible architectures and variations.

Therefore, we recommend exploring other transformer architectures such as BERT and GPT-2 to determine their effectiveness in emotion detection. It would also be interesting to investigate how different types of text data (e.g., social media posts vs. news articles) affect model performance and whether there are specific types of text data that are more difficult for models to classify accurately.

One possible reason why twitter-roberta-base-emotion outperformed other models might be it was pre-trained on a large corpus of text data and fine-tuned on a specific task. This allowed the model to learn general features of language that are useful for many tasks as well as task-specific features that are important for emotion classification.

In conclusion, we found that twitter-roberta-base-emotion outperformed other models such as LSTM with glove embedding or distilbert-base-uncased. We also found that preprocessing the data did not improve model performance and may have even led to a loss of accuracy. Our study highlights the importance of fine-tuning large datasets and not preprocessing data before passing it to transformers.

**Conflicts of Interest:** There is no conflict of interest across the authors.